\documentclass[runningheads]{llncs}
\usepackage{graphicx}

\usepackage{tikz}
\usepackage{comment}
\usepackage{amsmath,amssymb} 
\usepackage{color}

\usepackage[accsupp]{axessibility}  

\usepackage{booktabs}
\usepackage{multirow}

\usepackage{dsfont}
\usepackage{mathrsfs}
\usepackage{array}

\begin{document}
	\pagestyle{headings}
	\mainmatter
	\def\ECCVSubNumber{790}  
	
	\title{Learn-to-Decompose: Cascaded Decomposition Network for Cross-Domain Few-Shot Facial Expression Recognition} 

	\titlerunning{Learn-to-Decompose: CDNet for CD-FSL Facial Expression Recognition}
	%
	\author{Xinyi Zou\inst{1}\and
		Yan Yan\inst{1}\thanks{Corresponding author.}\and
		Jing-Hao Xue\inst{2}\and
		Si Chen\inst{3} \and
		Hanzi Wang\inst{1}
	}
	\authorrunning{Zou et al.}
	%
	\institute{Xiamen University, China \and
		University College London, UK \and
		Xiamen University of Technology, China\\
		\email{\{yanyan, hanzi.wang\}@xmu.edu.cn, zouxinyi0625@gmail.com}\\
		\email{jinghao.xue@ucl.ac.uk, chensi@xmut.edu.cn}
	}
	\maketitle
	
	\begin{abstract}
		Most existing compound facial expression recognition (FER) methods rely on large-scale labeled compound expression data for training. However, collecting such data is labor-intensive and time-consuming. In this paper, we address the compound FER task in the \emph{cross-domain few-shot learning} (FSL) setting, which requires only a few samples of compound expressions in the target domain. Specifically, we propose a novel cascaded decomposition network (CDNet), which cascades several learn-to-decompose modules with shared parameters based on a sequential decomposition mechanism, to obtain a transferable feature space. To alleviate the overfitting problem caused by limited base classes in our task, a partial regularization strategy is designed to effectively exploit the best of both episodic training and batch training. By training across similar tasks on multiple basic expression datasets, CDNet learns the ability of  \emph{learn-to-decompose} that can be easily adapted to identify unseen compound expressions.  Extensive experiments on both in-the-lab and in-the-wild compound expression datasets demonstrate the superiority of our proposed CDNet against several state-of-the-art FSL methods. Code is available at: https://github.com/zouxinyi0625/CDNet.
		
	\end{abstract}

	\section{Introduction}
	
	
	
	Automatic facial expression recognition (FER) is an important computer vision task with a variety of applications, such as mental assessment, driver fatigue surveillance, and interaction entertainment \cite{li2020deep}.

	The conventional FER task \cite{zeng2018facial,wang2020suppressing,ruan2021feature} aims to classify the input facial images into several basic expression categories (including anger, disgust, fear, happiness, sadness, surprise, contempt, and neutral). Unfortunately, the above basic expression categories cannot comprehensively describe the diversity and complexity of human emotions in practical scenarios. Later, Du \emph{et al.} \cite{du2014compound} further define compound expression categories. 
	Typically, compound expressions are much more fine-grained and difficult to be identified than basic expressions.
	Most existing work on compound FER \cite{slimani2019compound,guo2017multi} depends heavily on large-scale labeled {compound expression} data. However, annotating such data is time-consuming since the differences between compound expressions are subtle.

	
	To avoid expensive annotations, few-shot learning (FSL) 
	has emerged as a promising learning scheme. Very recently, Zou \emph{et al.} \cite{zou2022facial} first study cross-domain FSL for compound FER. 
	Our paper 
	is under the same problem setting as \cite{zou2022facial}, where easily-accessible basic expression datasets 
	and the compound expression dataset 
	are used for training and testing, respectively.
	However, unlike \cite{zou2022facial}, we address this problem from the perspective of feature decomposition and develop a novel regularization strategy for better performance.
	
	\begin{figure}[t]
		\centering
		\includegraphics[height=3.45cm]{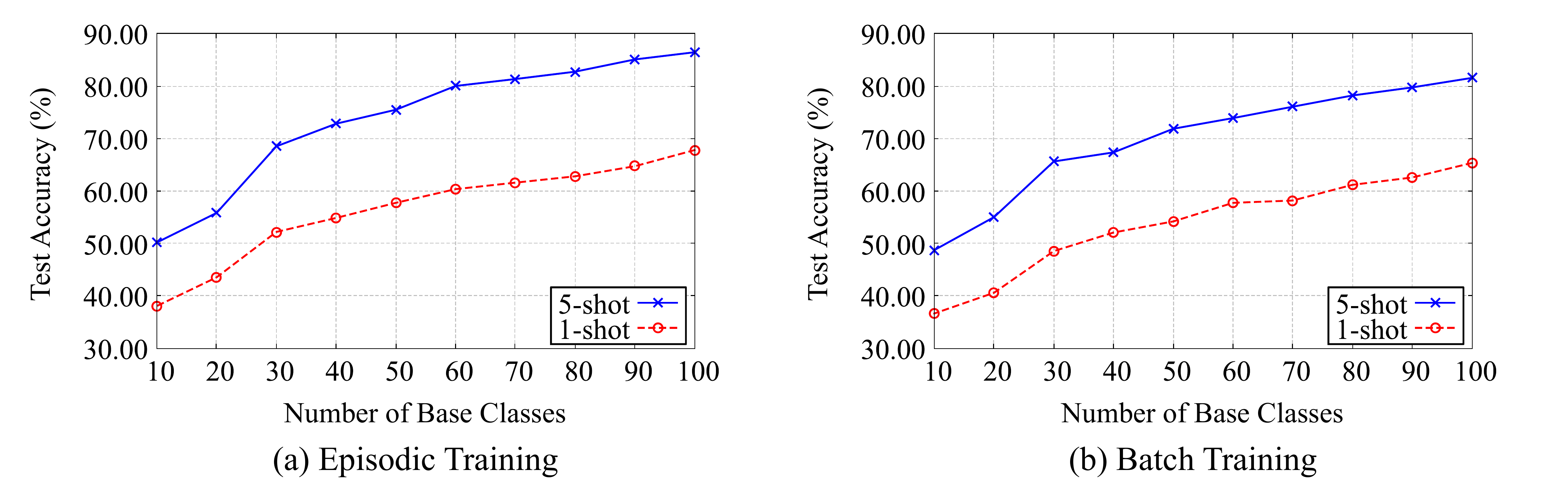}
		\caption{~Test accuracy obtained by (a) an episodic training-based FSL method (ProtoNet \cite{snell2017prototypical}) and (b) a batch training-based FSL method (BASELINE \cite{chen2019closer}) with the different numbers of base classes on  {a widely used FSL benchmark (CUB \cite{wah2011caltech})}.}
		\label{fig:motivation}
	\end{figure}
	%
	%
	%

	A key issue of FSL is how to obtain a transferable feature space.
	Two popular paradigms to learn such a space are episodic training and batch training.
	Episodic training-based methods \cite{liu2021learning,snell2017prototypical,vinyals2016matching}  construct few-shot tasks to learn knowledge transfer across similar tasks. Batch training-based methods \cite{afrasiyabi2020associative,chen2019closer,tian2020rethinking} learn a classification model to capture the information of all base classes. 
	Note that, different from popular FSL tasks {(such as image classification)}, which leverage a large number of base classes for training, our FER task involves {only} limited base classes (i.e.,
	the number of  classes in basic expression datasets is small). Hence, for episodic training-based methods, the sampled few-shot tasks are highly overlapped, leading to overfitting to the seen tasks. On the other hand, for batch training-based methods, the global view knowledge learned from limited base classes fails to be transferred to a novel task due to their inferior meta-learning ability.
	As shown in Fig.~\ref{fig:motivation}, the performance of exiting FSL methods drops substantially with the decreasing number of base classes.
	

	To alleviate the overfitting problem caused by limited base classes,  one reasonable way is to impose the batch training-based regularization to episode training. 
	Although some methods \cite{chen2020diversity,zhou2021binocular} employ the batch training as an auxiliary task, they do not work on the case of limited base classes. EGS-Net \cite{zou2022facial} applies full regularization of batch training to facilitate the training. However, the inferior meta-learning ability of batch training will unavoidably affect episodic training when full regularization is used. Therefore, it is significant to investigate how to exploit the best of both episodic training and batch training under limited base classes.  

	In this paper, we propose a novel cascaded decomposition network (CDNet), which cascades several learn-to-decompose (LD) modules with shared parameters in a sequential way, 
	for compound FER.  {Our method} is {inspired} by the RGB decomposition, where colors are represented by combinations of  R, G, and B values. Once  {a} model learns the ability of  RGB decomposition from existing colors, it can be easily adapted to infer new colors.  {In} the same spirit,
	we  {aim to} represent facial expressions as weighted  {combinations} of   expression prototypes.
	The expression prototypes encode the underlying generic knowledge across expressions while their weights characterize adaptive embeddings of one expression.
	{Building} on our cascaded decomposition,
	a partial regularization strategy is designed to effectively integrate episodic training and batch training. 
	
	Specifically,  {an} LD module, consisting of a decomposition block and a weighting block, learns an expression prototype and its corresponding weight, respectively. Based on the sequential decomposition mechanism, CDNet cascades LD modules with shared parameters to obtain weighted expression prototypes and reconstruct the expression feature. 
	During the episodic training of CDNet, we further leverage the batch training-based pre-trained model to regularize {only} the decomposition block (instead of full regularization on the whole LD module).
	{In this way}, a generic LD module  {can be} effectively learned under the supervision of the pre-trained model holding the global view of all base classes. 
	By training across similar tasks with our  {designed} partial regularization, CDNet is {enabled} to have the ability of learn-to-decompose that can be adapted to a novel task.
	
	In summary, our main contributions are given as follows:
	
	\begin{itemize}
		\item We propose a novel CDNet for compound FER in the cross-domain FSL setting.  {An} LD module is repeatedly exploited to extract the domain-agnostic expression feature from multi-source domains via a sequential decomposition mechanism. Based on the trained model, we can easily construct a  transferable feature space for the compound FER task in the target domain.
		
		\item We develop a partial regularization strategy to
		combine the benefits of episodic training and batch training. 
		Such a way can greatly alleviate the overfitting problem caused by limited base classes in our task and simultaneously maintain the meta-learning ability of the whole model.
		
		\item We perform extensive ablation studies to validate the importance of each component of CDNet. Experimental results show that CDNet performs favorably against state-of-the-art FSL methods for compound FER.
	\end{itemize}

	\section{Related Work}
	
	\subsection{Facial Expression Recognition}
	
	\noindent\textbf{Basic FER}
	\ Based on the Ekman and Friesen's study \cite{ekman1971constants}, the conventional FER task classifies an input facial image into one of the basic expression categories. In this paper, we refer to  the conventional  FER task as the basic FER. A large number of basic FER methods \cite{ruan2020deep,ruan2021feature,wang2019identity} have been proposed to extract discriminative expression features. 
	Recently, Ruan \emph{et al.} \cite{ruan2021feature} introduce a decomposition module to model  action-aware latent features for basic FER.
	Different from the parallel design in \cite{ruan2021feature}, we aim to learn a generic LD module by developing a sequential decomposition mechanism and a partial regularization strategy, enabling our model to obtain transferable features {in the FSL setting}.
	
	
	
	\noindent\textbf{Compound FER}
	\ Li \emph{et al.} \cite{li2019separate} design a novel separate loss to maximize the intra-class similarity and minimize the inter-class similarity for compound FER. Zhang \emph{et al.} \cite{zhang2020two} propose a coarse-to-fine two-stage strategy to enhance the robustness of the learned feature. Note that the above methods often require large-scale labeled compound expression data for training. Unfortunately, annotating these data is expensive and requires the professional guidance of psychology.
	
	
	\noindent\textbf{Few-Shot FER}
	\ Ciubotaru \emph{et al.} \cite{ciubotaru2019revisiting} revisit  popular FSL methods on basic FER. Recently, Zhu \emph{et al.} \cite{zhu2022convolutional} construct a convolutional relation
	network (CRN) to identify novel basic expression categories. The {work} most relevant  to ours is EGS-Net \cite{zou2022facial},
	which first investigates compound FER in the cross-domain FSL setting.
	Different from \cite{zou2022facial}, we propose CDNet  to obtain multiple weighted expression prototypes and reconstruct a transferable expression feature space. 
	
	\subsection{Few-Shot Learning}
	
	\noindent\textbf{Episodic Training-Based FSL}
	\ Based on what a model is expected to meta-learn \cite{lu2020learning}, various episodic training-based methods (such as {learn-to-measure \cite{snell2017prototypical,sung2018learning},} learn-to-fine-tune \cite{finn2017model,lee2018gradient}, {and} learn-to-parameterize \cite{gidaris2018dynamic,zhao2018dynamic}) 
	have been developed. In this paper,
	inspired by the fact that different colors can be reconstructed by a linear combination of R, G,
	and B values,
	we represent expression categories by combinations of
	{expression prototypes}. Technically, we {cascade several LD modules with shared parameters in a sequential decomposition way} to enforce the model to have the ability of learn-to-decompose by episodic training.
	
	\noindent\textbf{Batch Training-Based FSL}
	\ Recently, Chen \emph{et al.} \cite{chen2019closer} reveal that a simple baseline with a cosine classifier can  achieve surprisingly competing results.
	{Tian \emph{et al.} \cite{tian2020rethinking}  boost the performance with self-distillation.}
	Afrasiyabi \emph{et al.} \cite{afrasiyabi2020associative} develop a strong baseline
	with a novel angular margin loss and an early-stopping strategy.
	However, the meta-learning ability of batch training-based methods is
	inferior,
	especially when the number of base classes is limited  {as} in our task.

	\noindent\textbf{Hybrid FSL}
	\ A simple way to combine episodic training and batch training is Meta-Baseline \cite{chen2021meta}, which pre-trains the model by batch training and fine-tunes it by episodic training. Chen \emph{et al.} \cite{chen2020diversity} take the classification task as an auxiliary task to stabilize the training.
	Zhou \emph{et al.} \cite{zhou2021binocular} introduce binocular mutual learning (BML) to use the complementary information of the two paradigms.
	The above methods focus on popular FSL tasks with a large number of base classes for training. However, our FER task involves only limited base classes. As a result, these methods do not work well in our  {case}. In this paper, a partial regularization strategy is proposed to address the limited base classes problem by properly exploiting the advantages of  {the} two training paradigms. 

	\section{The Proposed Method}

	\subsection{Problem Definition}

	In this paper, as done in EGS-Net \cite{zou2022facial}, we consider the compound FER task in the cross-domain FSL setting, where only a few novel class samples are required to identify a compound expression category in the target domain.
	
	Given a labeled training set $\mathcal{D}_{train}$ (consisting of $C_{base}$ base classes), we aim to learn a model that can be well generalized to the test set $\mathcal{D}_{test}$ (consisting of $C_{novel}$ novel classes).
	In our setting, the base classes are the basic expression categories, while the novel classes are the compound expression categories.
	To enrich the diversity of the training set and handle the discrepancy between source and target domains, multiple easily-accessible basic expression datasets (i.e., the multi-source domains) are used.
	Note that the base  classes and novel classes are disjoint, and the number of base classes is limited {in our task.}
	
	\begin{figure}[!t]
		\centering
		\includegraphics[height=6.3cm]{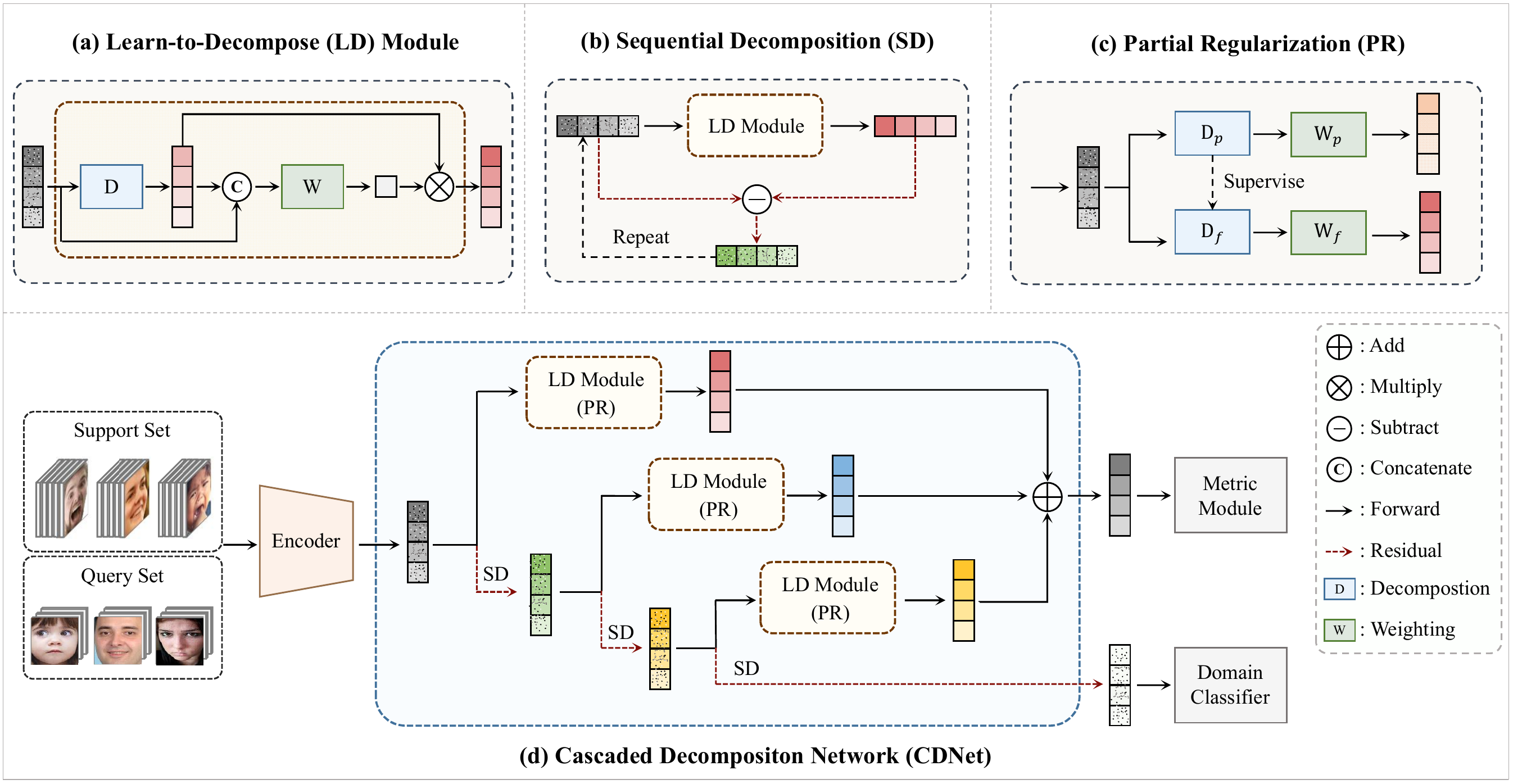}
		\caption{~{Overview of the proposed CDNet. It contains three main parts: (a) {an} LD module to extract an expression prototype and its corresponding weight, (b) a sequential decomposition mechanism to integrate shared LD modules, and (c) a partial regularization strategy to regularize the training. (d) CDNet is trained to obtain a transferable expression feature space that can be easily adapted to a novel compound FER task.}}
		\label{fig:overview}
	\end{figure}
	
	After training the model on $\mathcal{D}_{train}$, few-shot tasks are constructed on $\mathcal{D}_{test}$ to evaluate the performance of the learned model in the target domain. The goal of a few-shot task is to classify the query images with the reference of the support images. 
	Each few-shot task (an $N$-way $K$-shot task) samples $N$ classes from the $C_{novel}$ classes, and each class contains $K$ labeled support samples and $Q$ unlabeled query samples.
	In this paper, following the representative FSL method \cite{snell2017prototypical}, a query image is simply assigned to its nearest class in the learned expression feature space. {Hence, the key question of our task is how to construct a transferable expression feature space given limited base classes in $\mathcal{D}_{train}$.}
	
	\subsection{Overview}
	
	An overview of the proposed CDNet is shown in Fig.~\ref{fig:overview}.
	CDNet is composed of several learn-to-decompose (LD) modules with shared parameters.
	Each LD module (Fig.~\ref{fig:overview}(a)) consists of a decomposition block to extract an expression prototype and a weighting block to output the corresponding weight.
	Specifically, given an input feature from the encoder, {an} LD module first generates a weighted prototype, while the residual feature (obtained by subtracting the weighted prototype from the input feature) is then fed into the subsequent LD module.
	Based on a sequential decomposition mechanism (Fig.~\ref{fig:overview}(b)),  the LD module is repeatedly used to extract weighted prototypes progressively. 
	Finally, the expression feature, which is obtained by combining all the weighted prototypes, models the domain-agnostic expression information.  Meanwhile, the final residual feature
	captures the domain-specific information by
	identifying the input domain. As a consequence, CDNet can be well generalized to the unseen target domain.
	
	
	The training of CDNet involves two stages. {In the first stage, CDNet is pre-trained {in a batch training manner} to capture the global view information of all base classes on the whole $\mathcal{D}_{train}$.} In the second stage, CDNet is fine-tuned in an episode training manner under the regularization of the pre-trained model. To alleviate the overfitting problem caused by limited base classes, a novel partial regularization strategy (Fig.~\ref{fig:overview}(c)) is developed to  {regularize} the training in this stage. 
	By training across similar tasks with proper regularization, CDNet learns the ability of learn-to-decompose that can be easily adapted to a novel task.

	
	\subsection{Cascaded Decomposition Network (CDNet)}
	
	{CDNet consists of shared LD modules to reconstruct the {transferable} expression feature based on a novel sequential decomposition mechanism. In the following, we {will} elaborate the LD module and the sequential decomposition mechanism.}
	
	

	\noindent\textbf{LD Module}
	\ The LD module consists of a decomposition block and a weighting block.
	Specifically,  given an input feature $\boldsymbol{x}\in \mathbb{R}^{d}$, where $d$ denotes the dimension of the given feature, $\boldsymbol{x}$ is first fed into a decomposition block $\mathrm{D}(\cdot)$ to extract an expression prototype. The decomposition block includes a transform matrix $\mathrm{P}\in \mathbb{R}^{d\times d}$ and a PReLU activation layer $\mathrm{\sigma}(\cdot)$. 
	Mathematically, the expression prototype $\boldsymbol{p}$ is computed as
	\begin{equation}
		\begin{split}
			\label{eq:1}
			\boldsymbol{p} =~ &\mathrm{D}(\boldsymbol{x})\\
			=~&\mathrm{\sigma}(\mathrm{P}(\boldsymbol{x})).
		\end{split}
	\end{equation}
	~~~~Then, the extracted prototype $\boldsymbol{p}$ and the input feature $\boldsymbol{x}$ are concatenated, and they are fed into a weighting block $\mathrm{W}(\cdot)$ to compute the corresponding weight $\alpha=\mathrm{W}([\boldsymbol{x},\boldsymbol{p}])$, where $[\cdot,\cdot]$ represents the concatenation operation. The weighting block contains a three-layer perceptron. 
	{Finally,} the output of the LD module is the weighted prototype $\boldsymbol{f}$, which can be written as
	\begin{equation}
		\begin{split}
			\label{eq:wp}
			\boldsymbol{f}=~&\alpha\cdot \boldsymbol{p} \\
			=~&\mathrm{W}([\boldsymbol{x},\mathrm{D}(\boldsymbol{x})]) \cdot \mathrm{D}(\boldsymbol{x}).
		\end{split}
	\end{equation}
	\begin{figure}[t]
		\centering
		
		\includegraphics[height=2.6cm]{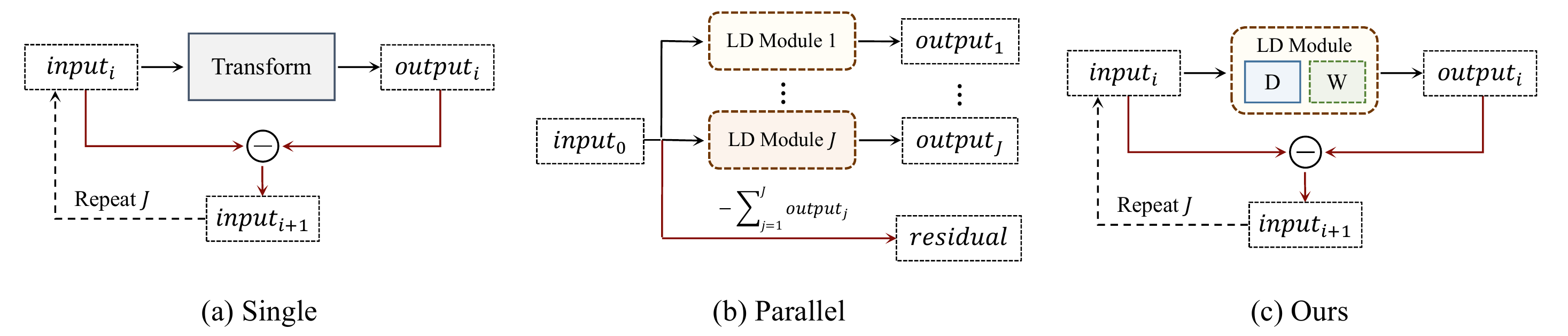}
		\caption{~Variants of the proposed CDNet. (a) A single transformation module which directly obtains a weighted prototype. (b) The parallel mechanism which simultaneously integrates multiple LD modules. (c) Our method which cascades the two-block LD modules with shared parameters based on a sequential decomposition mechanism.}
		\label{fig:varitions}
	\end{figure}

	Note that, instead of using a single transformation module (Fig. \ref{fig:varitions}(a)), we use two sequential blocks to obtain a weighted prototype {(Fig.~\ref{fig:varitions}(c))}. Such a way can better learn the expression prototype and enable us to impose the regularization only on the decomposition block. This is helpful to transfer the global view knowledge of all base classes by batch training and simultaneously preserve the meta-learning ability of the whole model by episodic training. 
	%
	
	\noindent\textbf{Sequential Decomposition Mechanism}
	\ A single weighted prototype cannot comprehensively provide the representation of the expression feature. The residual feature also contains the discriminative information for expression classification. Therefore, multiple weighted prototypes are desirable.
	
	As shown in Fig.~\ref{fig:varitions}(b), a straightforward way to learn multiple weighted prototypes is to use multiple {different} LD modules in a parallel way, where a separate LD module is required to compute a weighted prototype. However, multiple LD modules with different parameters are  not  suitable to be adapted to a novel task, since the {inference ability} of  over-parameterized LD modules is weak for each meta-task {involving only few training samples}.
	Here, we propose a sequential decomposition mechanism (Fig.~\ref{fig:varitions}(c)), which cascades the LD modules with shared parameters to obtain multiple weighted prototypes progressively. In this way, a generic LD module can be learned {by repeatedly using it for training}. 
	
	Specifically, the original feature $\boldsymbol{x}_0$ from the feature encoder is fed into the LD module to obtain the first weighted prototype $\boldsymbol{f}_{1}$ by Eq.~(\ref{eq:wp}), that is, $\boldsymbol{f}_1=\alpha_{1}\cdot \boldsymbol{p}_{1}$, where $\boldsymbol{p}_{1}$ and $\alpha_{1}$ denote the first expression prototype and its corresponding weight, respectively. Then, the residual feature is computed by subtracting the weighted prototype $\boldsymbol{f}_{1}$ from the original feature $\boldsymbol{x}_0$, and is further fed into the same LD module to obtain the second weighted prototype $\boldsymbol{f}_{2}$. The process is repeated several times to get the weighted prototypes progressively. For the $i$-th LD module, its input $\boldsymbol{input}_{i}$ and output $\boldsymbol{output}_i$ {are} defined as
	
	\begin{equation}
		\label{eq:input}
		\boldsymbol{input}_{i}=\left\{\begin{matrix}
			\boldsymbol{x}_0 & ,i=1 \\
			\boldsymbol{x}_0-\sum_{k=1}^{i-1}\alpha_{k} \boldsymbol{p}_{k} & ,i>1,
		\end{matrix}\right.
	\end{equation}
	
	\begin{equation}
		\boldsymbol{output}_{i}=\alpha_{i}\boldsymbol{p}_{i},
	\end{equation}
	where $\boldsymbol{x}_0$ is the original feature. $\boldsymbol{p}_{i}=\mathrm{D}(\boldsymbol{input}_{i})$ and  $\alpha_{i}=\mathrm{W}([\boldsymbol{input}_{i},\boldsymbol{p}_i])$ respectively denote the $i$-th expression prototype and its corresponding weight. 
	
	Finally, the expression feature is reconstructed by combining all the weighted prototypes to recognize the expression. At the same time, the residual feature  is used to identify the input domain, encoding the domain-specific information.
	Mathematically, the output of CDNet that cascades $J$ LD modules is
	\begin{equation}
		\label{eq:output}
		\mathrm{CDNet}(\boldsymbol{x}_0)=\left\{\begin{matrix}
			\boldsymbol{r}_{e}=&\sum_{k=1}^{J}\alpha_{k} \boldsymbol{p}_{k}\\
			\boldsymbol{r}_{d}=&\boldsymbol{x}_0-\boldsymbol{r}_{e},
		\end{matrix}\right.
	\end{equation}
	where $\boldsymbol{r}_{e}$ and $\boldsymbol{r}_{d}$ represent the domain-agnostic expression feature and the domain-specific residual feature, respectively. 
	
	By disentangling the domain information from the given feature,  we are able to extract a domain-agnostic expression feature. Therefore, we overcome the discrepancy between source and target domains, thus facilitating {the recognition of} compound expressions in the target domain.

	
	\subsection{Training Process}
	
	{CDNet is first pre-trained by batch training, and then fine-tuned by episodic training. In particular, a partial regularization strategy is designed to take full advantage of the two training paradigms  under the limited base class setting.}

	\noindent\textbf{Pre-training Stage}
	\ In this stage, CDNet is pre-trained in the batch training manner to obtain initial parameters.
	Moreover, the pre-trained decomposition block is used to regularize the fine-tuning stage.
	For each iteration, we sample a batch of data from a randomly selected source domain. A sample is denoted as  $(\boldsymbol{I},y_{e}, y_{d})$, where $\boldsymbol{I}$, $y_{e}$, and $y_{d}$ are the image, the expression label, and the domain label of the sample, respectively. $\boldsymbol{I}$ is fed into the feature encoder to obtain the original feature $\boldsymbol{x}_0$, which is further passed through CDNet to extract the  expression feature $\boldsymbol{r}_{e}$ and the residual feature $\boldsymbol{r}_{d}$, according to Eq.~(\ref{eq:output}).
	
	On one hand,  $\boldsymbol{r}_{e}$ is used to make prediction of the expression category, and the classification loss $\mathcal{L}_{cls}^{p}$ is defined by the popular cross-entropy loss:
	\begin{equation} \mathcal{L}_{cls}^{p}=-\sum_{c=1}^{C_{e}}\mathds{1}_{[c=y_{e}]}\mathrm{log}(\mathrm{F}_{e}(\boldsymbol{r}_{e})),
	\end{equation}
	where $\mathrm{F}_{e}(\cdot)$ is a linear expression classifier (i.e., a fully-connected layer), $C_{e}$ is the number of  basic expression categories, and $\mathds{1}_{[c=y_{e}]}$ equals to 1 when $c=y_{e}$, and 0 otherwise.

	On the other hand,  $\boldsymbol{r}_{d}$ is used to identify the input domain, so that  the learned expression feature
	is domain-agnostic and can be better adapted to the unseen target domain. The domain classification loss $\mathcal{L}_{d}^{p}$ is also the cross-entropy loss between the predicted results $\hat{y}_{d}$ and the ground-truth $y_{d}$:
	\begin{equation}
		\label{eq:domain} \mathcal{L}_{d}^{p}=-\sum_{c=1}^{C_{d}}\mathds{1}_{[c=y_{d}]}\mathrm{log}(\mathrm{F}_{d}(\boldsymbol{r}_{d})),
	\end{equation}
	where $\mathrm{F}_{d}(\cdot)$ is a domain classifier (i.e., a  two-layer perceptron) and $C_d$ is the number of source domain categories.
	
	Therefore, the total loss in this stage is the joint loss of $\mathcal{L}_{cls}^{p}$ and  $\mathcal{L}_{d}^{p}$:
	\begin{equation}
		\label{eq:weight1}
		\mathcal{L}_{p}=\mathcal{L}_{cls}^{p}+\lambda_{d}^p \mathcal{L}_{d}^p,
	\end{equation}
	where  $\lambda_{d}^p$ is the {balance} weight in the pre-training stage. 

	\noindent\textbf{Fine-tuning Stage}
	\ In this stage, CDNet is fine-tuned to learn transferable knowledge across similar tasks by episodic training.
	For each episode, an FSL task of $N$ classes is sampled on a randomly selected source domain. The support set $\mathbb{S}=\{\mathbb{I}_{s}, \mathbb{Y}_{s}\}$ and the query set $\mathbb{Q}=\{\mathbb{I}_{q}, \mathbb{Y}_{q}\}$ are constructed with $K$ support samples and $Q$ query samples of each class, where $\mathbb{I}_{*}$ and $\mathbb{Y}_{*}$ denote the image set and the corresponding label set. All the images are subsequently fed into the feature encoder and CDNet to obtain the expression feature and the domain-specific residual feature. 
	The expression feature and the domain-specific residual feature of an image in $\mathbb{S}$ ($\mathbb{Q}$) are denoted as $\boldsymbol{r}_{e}^{s}$ ($\boldsymbol{r}_{e}^{q}$) and  $\boldsymbol{r}_{d}^{s}$ ($\boldsymbol{r}_{d}^{q}$), respectively.

	The expression feature obtained by CDNet is used for expression classification. Following the representative ProtoNet \cite{snell2017prototypical}, each query image is assigned to its nearest support class center in the learned feature space. The expression classification loss of a query image is 
	\begin{equation} \mathcal{L}_{cls}^{f}=-\sum_{n=1}^{N}\mathds{1}_{[n=y_{q}]}\mathrm{log}(\mathrm{softmax}(-\mathcal{M}(\boldsymbol{r}_{e}^{q},R_{n}))),
	\end{equation}
	where $\boldsymbol{r}_{e}^{q}$ and $y_q$ are the expression feature and the expression label of the query image, respectively.
	$R_{n}=\frac{1}{K}\sum_{k=1}^{K}(\boldsymbol{r}_{e}^{s})_k^n$ represents the center of class $n$, where
	$(\boldsymbol{r}_{e}^{s})_k^n$ is the expression feature of the $k$-th image in class $n$ in the support set.
	$\mathcal{M}(\cdot)$ denotes the metric module (the Euclidean distance is used). $N$ is the number of sampled classes.
	$\mathrm{softmax}(\cdot)$ denotes the softmax function.
	
	The domain classification loss $\mathcal{L}_{d}^{f}$ is  similar to Eq.~(\ref{eq:domain}), where all the support samples and query samples are used to compute the domain classification loss.
	
	Finally, instead of applying full regularization to the whole LD module, we leverage
	partial regularization, which imposes regularization only on the decomposition block.
	{In this way, the decomposition block learns the generic
		expression prototype while the weighting block learns adaptive weights without being affected by batch training.}
	Mathematically, a regularization loss is 
	{designed to enforce} the output of the decomposition block in the fine-tuning stage to be close to that of the pre-trained decomposition block.
	That is,
	\begin{equation} \mathcal{L}_{r}^{f}=\sum_{i=1}^{J}||\mathrm{D}_p(\boldsymbol{input}_{i})-\mathrm{D}_f(\boldsymbol{input}_{i})||_{2}^{2},
	\end{equation}
	where $\mathrm{D}_p(\cdot)$ and $\mathrm{D}_f(\cdot)$ are the decomposition blocks in the pre-training stage and the fine-tuning stage, respectively. $J$ is the number of cascaded LD modules. 

	Overall, the total loss of the fine-tuning stage is formulated as 
	\begin{equation}
		\label{eq:weight2}
		\mathcal{L}_{f}=\mathcal{L}_{cls}^{f}+\lambda_{d}^{f} \mathcal{L}_{d}^{f}+\lambda_{r}^{f} \mathcal{L}_{r}^{f},
	\end{equation}
	where $\lambda_{d}^f$ and $\lambda_{r}^f$ are the {balance} weights in the fine-tuning stage. 

	\section{Experiments}

	\subsection{Datasets}
	
	Our CDNet is trained on multiple basic expression datasets and tested on the compound expression dataset.
	
	\noindent\textbf{Basic Expression Datasets}
	We use five basic expression datasets, including three in-the-lab datasets (CK+ \cite{lucey2010extended}, MMI \cite{pantic2005web}, and Oulu-CASIA \cite{zhao2011facial})), and two in-the-wild datasets (the basic expression subset of RAF-DB \cite{li2017reliable} and SFEW \cite{dhall2011static}), to form the training set.
	\textbf{Compound Expression Datasets}
	Three compound expression datasets (CFEE \cite{du2014compound}, EmotioNet \cite{fabian2016emotionet}, and RAF-DB) are used to evaluate the performance of the learned model. To ensure the disjointness between base classes and novel classes, only the compound expression subsets of these datasets, denoted as CFEE\_C, EmotioNet\_C, and RAF\_C, are used for testing. The details of these datasets are given in supplementary materials.
	
	\subsection{Implementation Details}
	
	Our method is implemented with PyTorch. All the facial images are first aligned and resized to 256 $\times$ 256. Then, they are randomly cropped to 224 $\times$ 224, following by a random horizontal flip and color jitter as data augmentation for training. To address the different numbers of classes and the different label sequences of multi-source domains, we use a mapping function to unify the labels. The encoder of our CDNet is ResNet-18, which is pre-trained on the MS-Celeb-1M  dataset \cite{guo2016ms}. The number of the cascaded LD modules (whose detailed architecture is given in supplementary materials) $J$ is set to 3 by default.
	
	Our model is trained using the Adam algorithm with the learning rate of 0.0001, $\beta_1=0.500$, and $\beta_2=0.999$. For the pre-training stage, the model is optimized by 10,000 iterations and each iteration samples a mini-batch (with the batch size of 16) from a randomly selected source domain. The {balance} weight $\lambda_{d}^t$ in Eq.~(\ref{eq:weight1}) is empirically set to 1.0. For the fine-tuning stage, the model is fine-tuned by 100 episodes and each episode contains 100 few-shot tasks. For a few-shot task, we set the number of classes $N=5$, the number of support samples $K=1$ or 5, and the number of query samples $Q=16$ for each class. The  {balance} weights $\lambda_{d}^f$ and $\lambda_{r}^f$  in Eq.~(\ref{eq:weight2}) are set to 0.01 and 1.0, respectively.  {The influence of different balance weights is shown in supplementary materials.}
	The accuracy of {1,000} few-shot tasks sampled on the test set is used for evaluation.
	
	\subsection{Ablation Studies}

	The details of the compared variants of CDNet are shown in Table~\ref{tab:details}. ProtoNet \cite{snell2017prototypical}, which adopts ResNet-18 as the encoder, is taken as the baseline method.

	\label{sec:ab}
	
	\begin{table}[!t]
		\scriptsize
		\centering
		\caption{~The details of the baseline method and 6 variants of CDNet.}
		\label{tab:details}
		\setlength{\tabcolsep}{3.5mm}{
			\begin{tabular}{c|c|c|c}
				
				\toprule 
				Methods   & Decomposition Module & Decomposition Mechanism & Regularization \\
				\hline
				Baseline  &  $\times$  & $\times$         &    $\times$       \\
				Single    &  Single Transformation  &  Sequential          &   Full             \\
				Parallel  & LD   &     Parallel       &    Partial            \\
				Decompose & LD   & Sequential           &   $\times$          \\
				CDNet\_Full    &  LD  &  Sequential          &   Full             \\
				CDNet\_Fix    &  LD  &  Sequential          &   Partial (Fix)             \\
				CDNet (ours)     & LD   &  Sequential          &    Partial           \\
				\bottomrule 
			\end{tabular}
			
		}
		
	\end{table}

	\noindent\textbf{Influence of the LD Module}
	To validate the effectiveness of our proposed LD module, we replace the {LD} module with a  single transformation module (as shown in Fig.~\ref{fig:varitions}(a))
	to obtain the weighted prototype. We denote our method based on the single module as ``Single''.
	We also evaluate a variant of CDNet (denoted as ``CDNet\_Full''), which applies the pre-trained LD module to supervise the output of the {whole} LD module during the fine-tuning stage.
	The comparison results are shown in Table~\ref{tab:ab}.

	Compared with CDNet, the ``Single'' method performs worse since it can only impose full regularization on the whole transformation module. Moreover, 
	the ``CDNet\_Full'' method obtains better accuracy than the ``Single'' method. This indicates the superiority of obtaining the weighted prototype in our two-block LD module.
	By learning the expression prototype and its weight individually, we can
	fully exploit the {generic} information across basic expressions, which can facilitate  the extraction of a transferable expression feature.

	\noindent\textbf{Influence of the Sequential Decomposition Mechanism}
	In Table~\ref{tab:ab}, we compare our sequential decomposition mechanism with the parallel mechanism (Fig.~\ref{fig:varitions}(b)) that obtains multiple weighted prototypes using different LD modules. Our method based on the parallel mechanism is denoted as ``Parallel''.
	
	CDNet with our sequential decomposition mechanism outperforms that with the parallel mechanism. 
	The inference ability of the ``Parallel'' method is weak due to over-parameterized LD modules when only a small number of training samples are available. 
	These results show the effectiveness of using a generic LD module to obtain the weighted prototypes progressively {in} the FSL setting.
	
	\noindent\textbf{Influence of the Cascaded Decomposition Design}
	We evaluate the performance of our cascaded decomposition design (denoted as ``Decompose'') in Table~\ref{tab:ab}.
	We can see that the {``Decompose''} method performs better than the baseline method, and the improvements are more evident when using a batch training-based pre-trained model. 
	With the ability of learn-to-decompose, our method can extract transferable features for the novel compound FER, especially on the more difficult 1-shot classification task.
	
\begin{figure}[!t]
		
		\centering
		\includegraphics[height=2.4cm]{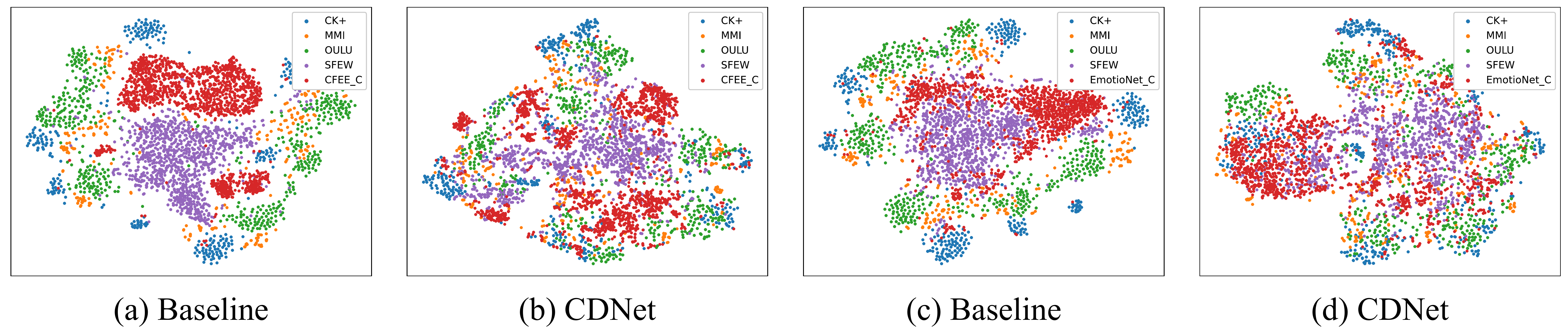}
		\caption{~t-SNE {visualization} of the extracted feature obtained by (a) the Baseline method and (b) CDNet on multi-source domains and a target domain (CFEE\_C). The target domain is marked in red. The domain discrepancy between multi-source domains and the target domain is reduced by our CDNet, facilitating our cross-domain FER task. (c-d) A similar pattern can be found for another target domain (EmotioNet\_C).}
		\label{fig:feature}
		
	\end{figure}

	Fig.~\ref{fig:feature} further illustrates t-SNE visualization results of the learned feature space by the baseline method and our CDNet. Clearly, the domain gap between the source and target domains is greatly reduced by our CDNet. 
	By disentangling the domain-specific information from the original feature, the learned expression feature is domain-agnostic, alleviating the domain discrepancy between the source and target domains.
	{The superiority of using a decomposition-based design is further illustrated in supplementary materials.}
	%

	\begin{table}[!t]
		\scriptsize
		\centering
		\caption{~ Influence of the key modules. Test accuracy (\%) of 5-way few shot classification tasks  with 95\% confidence intervals on  three different datasets. ``Pre'' indicates that the pre-trained model is used as an initialization in the second stage.}
		\label{tab:ab}
		\scalebox{0.99}{
			\begin{tabular}{c|c|cc|cc|cc}
				\toprule [1 pt]
				\multirow{2}{*}{\tiny{Pre}} & \multirow{2}{*}{Method} & \multicolumn{2}{c|}{CFEE\_C} & \multicolumn{2}{c|}{EmotioNet\_C} & \multicolumn{2}{c}{RAF\_C} \\
				&                         & 1-shot      & 5-shot      & 1-shot      & 5-shot      & 1-shot      & 5-shot                             \\

				\bottomrule [0.01 pt]
				\toprule [0.01 pt]
				
				$\times$ & Baseline        & 53.29 \tiny{$\pm$ 0.73}          & 66.60 \tiny{$\pm$ 0.60}        &  50.15 \tiny{$\pm$ 0.66}       & 60.04 \tiny{$\pm$ 0.56}          &39.12  \tiny{$\pm$   0.56}        &  58.41 \tiny{$\pm$   0.46}        \\
				$\times$ & Single      & 53.67 \tiny{$\pm$ 0.72}        & 67.25 \tiny{$\pm$ 0.61}         & 50.72 \tiny{$\pm$ 0.65}          & 60.62 \tiny{$\pm$ 0.56}          & 40.59 \tiny{$\pm$ 0.57}          & 59.40 \tiny{$\pm$ 0.46}          \\
				$\times$ & Parallel      & 53.83 \tiny{$\pm$ 0.73}        & 67.53 \tiny{$\pm$ 0.60}         & 51.84 \tiny{$\pm$ 0.67}          & 61.21 \tiny{$\pm$ 0.58}          &  41.11 \tiny{$\pm$ 0.57}          & 60.48 \tiny{$\pm$ 0.45}          \\
				$\times$ & Decompose      & 53.78 \tiny{$\pm$ 0.72}        & 67.60 \tiny{$\pm$ 0.61}         & 51.16 \tiny{$\pm$ 0.67}          & 60.75 \tiny{$\pm$ 0.58}          & 40.75 \tiny{$\pm$ 0.57}          & 59.69 \tiny{$\pm$ 0.44}          \\
				$\times$ & CDNet\_Full   & 54.23 \tiny{$\pm$ 0.70}         & 67.82 \tiny{$\pm$ 0.59}          & 51.72 \tiny{$\pm$ 0.68}          & 61.47 \tiny{$\pm$ 0.60}         & 41.51 \tiny{$\pm$ 0.59}         & 60.25 \tiny{$\pm$ 0.45}          \\
				$\times$ & 	CDNet   & 54.55 \tiny{$\pm$   0.71}          & 68.09 \tiny{$\pm$ 0.62}          & 52.76 \tiny{$\pm$ 0.67}         & 61.76 \tiny{$\pm$ 0.57}          & 42.02 \tiny{$\pm$ 0.58}          & 61.75 \tiny{$\pm$ 0.44}          \\
				
				\bottomrule [0.01 pt]
				\toprule [0.01 pt]
				
				\checkmark & Baseline     & 53.43 \tiny{$\pm$   0.71}          & 66.75 \tiny{$\pm$ 0.60}        & 51.31 \tiny{$\pm$ 0.67}           & 60.68 \tiny{$\pm$ 0.59}            & 42.30 \tiny{$\pm$ 0.58}         & 60.19 \tiny{$\pm$ 0.43}            \\
				\checkmark & Single     & 54.37 \tiny{$\pm$   0.71}          & 67.44 \tiny{$\pm$ 0.59}        & 53.13 \tiny{$\pm$ 0.65}           & 61.47 \tiny{$\pm$ 0.58}            & 43.22 \tiny{$\pm$ 0.58}         & 60.94 \tiny{$\pm$ 0.45}            \\
				\checkmark & Parallel   & 54.72 \tiny{$\pm$ 0.72}         & 67.66 \tiny{$\pm$ 0.59}          & 54.27 \tiny{$\pm$ 0.68}          & 61.74 \tiny{$\pm$ 0.59}         & 44.78 \tiny{$\pm$ 0.60}         & 61.29 \tiny{$\pm$ 0.43}          \\
				\checkmark & Decompose   & 55.17 \tiny{$\pm$ 0.71}         & 67.85 \tiny{$\pm$ 0.59}          & 53.01 \tiny{$\pm$ 0.68}          & 61.60 \tiny{$\pm$ 0.60}         & 43.78 \tiny{$\pm$ 0.59}         & 61.00 \tiny{$\pm$ 0.43}          \\
				\checkmark & CDNet\_Full   & 55.91 \tiny{$\pm$ 0.73}         & 68.37 \tiny{$\pm$ 0.60}          & 54.11 \tiny{$\pm$ 0.68}          & 61.99 \tiny{$\pm$ 0.61}         & 45.69 \tiny{$\pm$ 0.59}         & 61.59 \tiny{$\pm$ 0.42}          \\
				\checkmark & CDNet\_Fix   & 56.45 \tiny{$\pm$ 0.72}         & 68.54 \tiny{$\pm$ 0.61}          & 54.60 \tiny{$\pm$ 0.68}          & 62.50 \tiny{$\pm$ 0.60}         & 45.90 \tiny{$\pm$ 0.60}         & 62.12 \tiny{$\pm$ 0.43}          \\
				\checkmark & 	CDNet & \textbf{56.99 \tiny{$\pm$   0.73}} & \textbf{68.98 \tiny{$\pm$ 0.60}} & \textbf{55.16 \tiny{$\pm$ 0.67}} & \textbf{63.03 \tiny{$\pm$ 0.59}} & \textbf{46.07 \tiny{$\pm$ 0.59}} & \textbf{63.03 \tiny{$\pm$ 0.45}}\\
				\bottomrule [1 pt]
			\end{tabular}
		}
	\end{table}
	
	\noindent\textbf{Influence of Regularization}
	We also compare the CDNet  {methods} with and without our regularization {strategy} (denoted as ``CDNet'' and ``Decompose'', respectively) in Table \ref{tab:ab}. 
	We can see that the recognition accuracy is greatly improved when our regularization strategy is used.
	This validates the importance of imposing the {partial} regularization from a pre-trained block (which holds the global view) during episodic training. 

	Moreover, we compare the proposed partial regularization strategy with two variants (``CDNet\_Full'' and ``CDNet\_Fix''). 
	From {Table~\ref{tab:ab}}, the ``CDNet\_Full'' method achieves worse results than the other two regularization variants. This can be ascribed to the poor ``meta-learning'' ability of the pre-trained model, deteriorating the transfer ability of the fine-tuning stage.
	``CDNet\_Fix'' fixes the pre-trained decomposition block and only fine-tunes the weighting block to obtain adaptive weights.  {It} alleviates the overfitting problem with the fixed decomposition block, but the inference ability of the fixed decomposition block is still poor. Our method with the proposed partial regularization  {strategy} obtains the highest recognition accuracy among the three regularization variants.
	

	\noindent\textbf{Influence of the Pre-training Stage}
	In Table \ref{tab:ab}, all the variants with the pre-trained initialization achieve higher accuracy. 
	The improvements on CDNet are more evident than those on the baseline method. This shows the importance of preserving a global view  for feature decomposition  in our FER task. 

	%
	%
	%
	
\begin{figure}[!t]
		
		\centering
		\includegraphics[height=3.4cm]{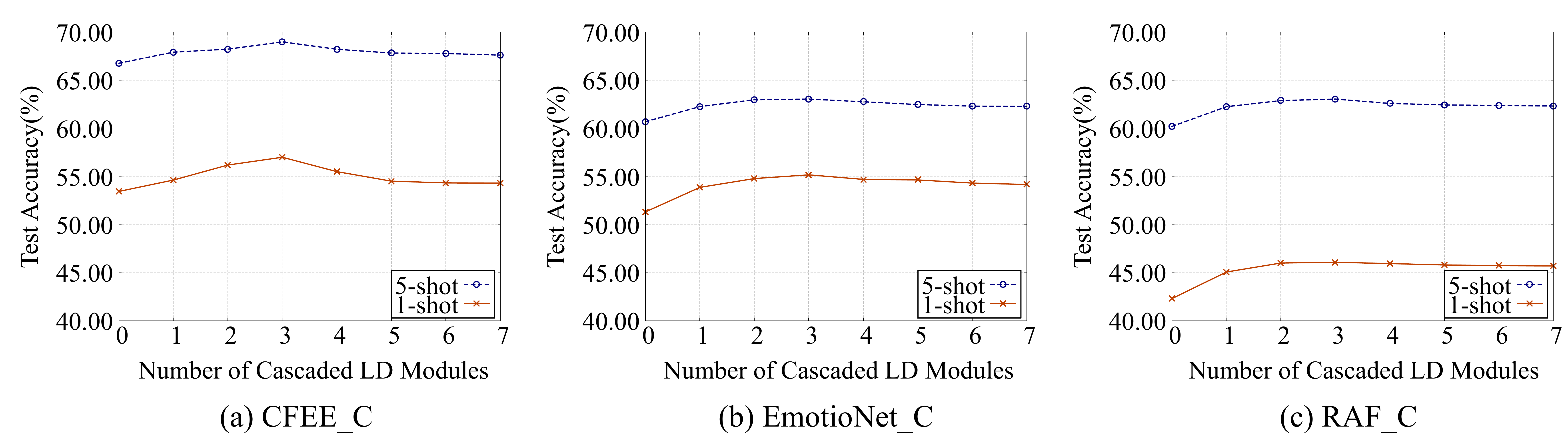}
		\caption{~Influence of the number of cascaded LD modules on three different datasets. The average accuracy of 5-way few-shot classification tasks is reported.}
		\label{fig:color}
		
	\end{figure}

	\noindent\textbf{Influence of the Number of Cascaded LD Modules}
	We also evaluate the performance with the different numbers of cascaded LD modules, as shown in Fig.~\ref{fig:color}. The 0-layer CDNet indicates the baseline method.
	Compared with the baseline method, CDNet with a single LD module can improve the accuracy by
	disentangling the expression feature and the domain-specific residual feature.
	The performance of CDNet is further boosted when two or three shared LD modules are cascaded to obtain weighted prototypes. This indicates that the residual feature from the first decomposition contains the expression information.
	However, when more cascaded LD modules are used, the accuracy decreases especially on the CFEE\_C dataset.
	In such a case, the residual feature is less informative, thus reducing the discrimination of the learned feature.
	Our method achieves the best performance when the number of cascaded LD modules is 3.

	%
	
	\subsection{Comparison with State-of-the-Art Methods}

	\begin{table}[!t]
		\scriptsize
		\caption{~Comparisons with state-of-the-art FSL methods. Test accuracy (\%) of 5-way few-shot classification tasks  with 95\% confidence intervals on three different datasets.} 
		\label{tab:sota}
		\centering
		
		\begin{tabular}{l|cc|cc|cc}
			\toprule [1 pt]
			\multirow{2}{*}{Method} & \multicolumn{2}{c|}{CFEE\_C}                             & \multicolumn{2}{c|}{EmotioNet\_C}                      & \multicolumn{2}{c}{RAF\_C}                            \\
			
			& 1-shot                      & 5-shot                    & 1-shot                    & 5-shot                    & 1-shot                    & 5-shot                    \\
			
			\bottomrule [0.01 pt]
			\multicolumn{7}{c}{(a) Episodic training-based FSL methods}  \\
			\toprule [0.01 pt]
			
			ProtoNet \cite{snell2017prototypical}        & 53.29 \tiny{$\pm$ 0.73}          & 66.60 \tiny{$\pm$ 0.60}        & 50.15 \tiny{$\pm$ 0.66}       & 60.04 \tiny{$\pm$ 0.56}          & 39.12 \tiny{$\pm$   0.56}        & 58.41 \tiny{$\pm$   0.46}        \\
			
			MatchingNet \cite{vinyals2016matching}        & 52.31 \tiny{$\pm$ 0.69}          & 62.24 \tiny{$\pm$ 0.61}        & 48.64 \tiny{$\pm$ 0.63}       & 54.19 \tiny{$\pm$ 0.58}          & 34.84 \tiny{$\pm$   0.54}        & 52.45 \tiny{$\pm$   0.44}        \\
			
			RelationNet \cite{sung2018learning}        & 50.58 \tiny{$\pm$ 0.68}          & 63.17 \tiny{$\pm$ 0.60}        & 48.33 \tiny{$\pm$ 0.68}       & 56.27 \tiny{$\pm$ 0.58}          & 36.18 \tiny{$\pm$   0.54}        & 53.45 \tiny{$\pm$   0.46}        \\
			
			GNN \cite{garcia2017few}        & 54.01 \tiny{$\pm$ 0.74}          & 64.26 \tiny{$\pm$ 0.63}        & 49.49 \tiny{$\pm$ 0.68}       & 58.67 \tiny{$\pm$ 0.59}          & 38.74 \tiny{$\pm$   0.56}        & 57.15 \tiny{$\pm$   0.47}        \\
			
			DSN \cite{simon2020adaptive}        & 49.61 \tiny{$\pm$ 0.73}          & 60.03 \tiny{$\pm$ 0.62}        & 48.25 \tiny{$\pm$ 0.68}       & 54.89 \tiny{$\pm$ 0.58}          & 40.09 \tiny{$\pm$   0.55}        & 52.49 \tiny{$\pm$   0.47}        \\
			
			InfoPatch \cite{liu2021learning}        & 54.19 \tiny{$\pm$ 0.67}          & 67.29 \tiny{$\pm$ 0.56}        & 48.14 \tiny{$\pm$ 0.61}       & 59.84 \tiny{$\pm$ 0.55}          & 41.02 \tiny{$\pm$   0.52}        & 57.98 \tiny{$\pm$   0.45}        \\
			
			\bottomrule [0.01 pt]
			\multicolumn{7}{c}{(b) Batch training-based FSL methods}  \\
			\toprule [0.01 pt]
			
			softmax \cite{chen2019closer}        & 54.32 \tiny{$\pm$ 0.73}          & 66.35 \tiny{$\pm$ 0.62}        & 51.60 \tiny{$\pm$ 0.68}       & 61.83 \tiny{$\pm$ 0.59}          & 42.16 \tiny{$\pm$   0.59}        & 58.57 \tiny{$\pm$   0.45}        \\
			
			cosmax \cite{chen2019closer}        & 54.97 \tiny{$\pm$ 0.71}          & 67.89 \tiny{$\pm$ 0.61}        & 50.87 \tiny{$\pm$ 0.65}       & 61.10 \tiny{$\pm$ 0.56}          & 40.87 \tiny{$\pm$ 0.56}        & 57.67 \tiny{$\pm$   0.46}        \\
			
			arcmax \cite{afrasiyabi2020associative}        & 55.29 \tiny{$\pm$ 0.71}          & 67.72 \tiny{$\pm$ 0.60}        & 50.73 \tiny{$\pm$ 0.65}       & 61.70 \tiny{$\pm$ 0.56}          & 41.28 \tiny{$\pm$   0.57}        & 57.94 \tiny{$\pm$   0.46}        \\
			
			rfs \cite{tian2020rethinking}        & 54.96 \tiny{$\pm$ 0.73}          & 65.71 \tiny{$\pm$ 0.61}        & 51.91 \tiny{$\pm$ 0.67}       & 61.94 \tiny{$\pm$ 0.57}          & 43.05 \tiny{$\pm$   0.59}        & 60.08 \tiny{$\pm$   0.46}        \\
			
			LR+DC \cite{yang2021free}        & 53.20 \tiny{$\pm$ 0.73}          & 64.18 \tiny{$\pm$ 0.66}        & 52.09 \tiny{$\pm$ 0.70}       & 60.12 \tiny{$\pm$ 0.58}          & 42.90 \tiny{$\pm$   0.60}        & 56.74 \tiny{$\pm$   0.46}        \\
			
			STARTUP \cite{phoo2020self}        & 54.89 \tiny{$\pm$ 0.72}          & 67.79 \tiny{$\pm$ 0.61}        & 52.61 \tiny{$\pm$ 0.69}       & 61.95 \tiny{$\pm$ 0.57}          & 43.97 \tiny{$\pm$   0.60}        & 59.14 \tiny{$\pm$   0.47}        \\
			
			\bottomrule [0.01 pt]
			\multicolumn{7}{c}{(c) Hybrid FSL methods}  \\
			\toprule [0.01 pt]
			
			Meta-Baseline \cite{chen2021meta}        & 55.17 \tiny{$\pm$ 0.74}          & 67.15 \tiny{$\pm$ 0.61}        & 52.36 \tiny{$\pm$ 0.67}       & 62.01 \tiny{$\pm$ 0.59}          & 43.54 \tiny{$\pm$   0.61}        & 61.59 \tiny{$\pm$   0.44}        \\

			OAT \cite{chen2020diversity}        & 54.28 \tiny{$\pm$ 0.75}          & 67.88 \tiny{$\pm$ 0.62}        & 52.92 \tiny{$\pm$ 0.66}       & 61.85 \tiny{$\pm$ 0.59}          & 42.75 \tiny{$\pm$   0.60}        & 60.41 \tiny{$\pm$   0.43}        \\
			
			BML \cite{zhou2021binocular}        & 52.42 \tiny{$\pm$ 0.71}          & 66.72 \tiny{$\pm$ 0.61}        & 51.31 \tiny{$\pm$ 0.66}       & 58.77 \tiny{$\pm$ 0.57}          & 41.91 \tiny{$\pm$   0.55}        & 59.72 \tiny{$\pm$   0.45}        \\
			
			EGS-Net \cite{zou2022facial}        & 56.65 \tiny{$\pm$ 0.73}          & 68.38 \tiny{$\pm$ 0.60}        & 51.62 \tiny{$\pm$ 0.66}       & 60.52 \tiny{$\pm$ 0.56}          & 44.07 \tiny{$\pm$   0.60}        & 61.90 \tiny{$\pm$   0.46}        \\
			
			\bottomrule [0.01 pt]
			\toprule [0.01 pt]

			\textbf{CDNet (ours)} & \textbf{56.99 \tiny{$\pm$   0.73}} & \textbf{68.98 \tiny{$\pm$ 0.60}} & \textbf{55.16 \tiny{$\pm$ 0.67}} & \textbf{63.03 \tiny{$\pm$ 0.59}} & \textbf{46.07 \tiny{$\pm$ 0.59}} & \textbf{63.03 \tiny{$\pm$ 0.45}}\\
			\bottomrule [1 pt]
		\end{tabular}
		
	\end{table}
	
	Table~\ref{tab:sota} shows the performance comparisons between our developed CDNet and several state-of-the-art FSL methods. 
	For a fair comparison, we report the results of the competing methods using the source codes provided by respective authors under the same settings as ours.
	
	{From Table~\ref{tab:sota}, the episodic training-based methods suffer from the overfitting problem caused by  highly overlapped sampled tasks under the limited base class setting in our task. 
		The batch training-based methods perform better than the episodic training-based methods since the extracted global view information 
		is not easily affected by  overlapped sampled tasks.
	}
	
	{Existing hybrid FSL methods combine the two training paradigms to facilitate the training. For example, Meta-Baseline \cite{chen2021meta} pre-trains the model by batch training and  {fine-tunes} it by episodic training. OAT \cite{chen2020diversity} proposes  {an} organized auxiliary task co-training method, which organizes the batch training and episodic training in an  orderly way to  {stabilize} the training process. BML \cite{zhou2021binocular} aggregates the complementary information of the batch training branch and the episodic training branch for the meta-test tasks. However, they do not focus on the limited base class setting.  {Very recently,} EGS-Net \cite{zou2022facial} uses a joint and alternate learning framework to alleviate the {problem of} limited base classes. Different from EGS-Net,  we incorporate a novel partial regularization strategy into CDNet to  {take full} advantage of the two training paradigms. Results show that our method outperforms the above methods for both the 1-shot and 5-shot few-shot classification tasks on all the compound expression datasets.}

	

	\section{Conclusions}
	
	In this paper, we address compound FER in the cross-domain FSL setting, which alleviates the burden of collecting large-scale labeled compound FER data. Based on {a proposed} sequential decomposition mechanism, we develop a novel CDNet consisting of cascaded LD modules with shared parameters to learn a transferable feature space.
	In particular, a partial regularization strategy is designed to take full advantage of episodic training and batch training. In this way, we prevent the model from overfitting to highly-overlapped seen tasks due to limited base classes and enable the trained model to have the ability of learn-to-decompose.  Experimental results  {show} the effectiveness of our proposed method {against several state-of-the-art FSL methods} on various compound FER datasets.
	
	\section*{Acknowledgement}
	This work was partly supported by the National Natural Science Foundation of
	China under Grants 62071404, U21A20514, and 61872307, by the Open
	Research Projects of Zhejiang Lab under Grant 2021KG0AB02, by the Natural
	Science Foundation of Fujian Province under Grant 2020J01001, and by the
	Youth Innovation Foundation of Xiamen City under Grant 3502Z20206046.
	
	%
	%
	\bibliographystyle{splncs04}
	\bibliography{ref}

\begin{thebibliography}{10}
\providecommand{\url}[1]{\texttt{#1}}
\providecommand{\urlprefix}{URL }
\providecommand{\doi}[1]{https://doi.org/#1}

\bibitem{afrasiyabi2020associative}
Afrasiyabi, A., Lalonde, J.F., Gagn{\'e}, C.: Associative alignment for
  few-shot image classification. In: European Conference on Computer Vision.
  pp. 18--35 (2020)

\bibitem{chen2020diversity}
Chen, M., Fang, Y., Wang, X., Luo, H., Geng, Y., Zhang, X., Huang, C., Liu, W.,
  Wang, B.: Diversity transfer network for few-shot learning. In: AAAI
  Conference on Artificial Intelligence. vol.~34, pp. 10559--10566 (2020)

\bibitem{chen2019closer}
Chen, W.Y., Liu, Y.C., Kira, Z., Wang, Y.C., Huang, J.B.: A closer look at
  few-shot classification. In: International Conference on Learning
  Representations (2019)

\bibitem{chen2021meta}
Chen, Y., Liu, Z., Xu, H., Darrell, T., Wang, X.: {Meta-Baseline}: Exploring
  simple meta-learning for few-shot learning. In: IEEE/CVF International
  Conference on Computer Vision. pp. 9062--9071 (2021)

\bibitem{ciubotaru2019revisiting}
Ciubotaru, A.N., Devos, A., Bozorgtabar, B., Thiran, J.P., Gabrani, M.:
  Revisiting few-shot learning for facial expression recognition. arXiv
  preprint arXiv:1912.02751  (2019)

\bibitem{dhall2011static}
Dhall, A., Goecke, R., Lucey, S., Gedeon, T.: Static facial expression analysis
  in tough conditions: Data, evaluation protocol and benchmark. In: IEEE
  International Conference on Computer Vision Workshops. pp. 2106--2112 (2011)

\bibitem{du2014compound}
Du, S., Tao, Y., Martinez, A.M.: Compound facial expressions of emotion.
  Proceedings of the National Academy of Sciences  \textbf{111}(15),
  E1454--E1462 (2014)

\bibitem{ekman1971constants}
Ekman, P., Friesen, W.V.: Constants across cultures in the face and emotion.
  Journal of Personality and Social Psychology  \textbf{17}(2),  124--129
  (1971)

\bibitem{fabian2016emotionet}
Fabian Benitez-Quiroz, C., Srinivasan, R., Martinez, A.M.: {EmotioNet}: An
  accurate, real-time algorithm for the automatic annotation of a million
  facial expressions in the wild. In: IEEE/CVF Conference on Computer Vision
  and Pattern Recognition. pp. 5562--5570 (2016)

\bibitem{finn2017model}
Finn, C., Abbeel, P., Levine, S.: Model-agnostic meta-learning for fast
  adaptation of deep networks. In: International Conference on Machine
  Learning. pp. 1126--1135 (2017)

\bibitem{garcia2017few}
Garcia, V., Bruna, J.: Few-shot learning with graph neural networks. In:
  International Conference on Learning Representations (2018)

\bibitem{gidaris2018dynamic}
Gidaris, S., Komodakis, N.: Dynamic few-shot visual learning without
  forgetting. In: IEEE/CVF Conference on Computer Vision and Pattern
  Recognition. pp. 4367--4375 (2018)

\bibitem{guo2017multi}
Guo, J., Zhou, S., Wu, J., Wan, J., Zhu, X., Lei, Z., Li, S.Z.: Multi-modality
  network with visual and geometrical information for micro emotion
  recognition. In: IEEE International Conference on Automatic Face and Gesture
  Recognition. pp. 814--819 (2017)

\bibitem{guo2016ms}
Guo, Y., Zhang, L., Hu, Y., He, X., Gao, J.: {Ms-Celeb-1M}: A dataset and
  benchmark for large-scale face recognition. In: European Conference on
  Computer Vision. pp. 87--102 (2016)

\bibitem{lee2018gradient}
Lee, Y., Choi, S.: Gradient-based meta-learning with learned layerwise metric
  and subspace. In: International Conference on Machine Learning. pp.
  2927--2936 (2018)

\bibitem{li2020deep}
Li, S., Deng, W.: Deep facial expression recognition: A survey. IEEE
  Transactions on Affective Computing  (2020)

\bibitem{li2017reliable}
Li, S., Deng, W., Du, J.: Reliable crowdsourcing and deep locality-preserving
  learning for expression recognition in the wild. In: IEEE/CVF Conference on
  Computer Vision and Pattern Recognition. pp. 2852--2861 (2017)

\bibitem{li2019separate}
Li, Y., Lu, Y., Li, J., Lu, G.: Separate loss for basic and compound facial
  expression recognition in the wild. In: Asian Conference on Machine Learning.
  pp. 897--911 (2019)

\bibitem{liu2021learning}
Liu, C., Fu, Y., Xu, C., Yang, S., Li, J., Wang, C., Zhang, L.: Learning a
  few-shot embedding model with contrastive learning. In: AAAI Conference on
  Artificial Intelligence. vol.~35, pp. 8635--8643 (2021)

\bibitem{lu2020learning}
Lu, J., Gong, P., Ye, J., Zhang, C.: Learning from very few samples: A survey.
  arXiv preprint arXiv:2009.02653  (2020)

\bibitem{lucey2010extended}
Lucey, P., Cohn, J.F., Kanade, T., Saragih, J., Ambadar, Z., Matthews, I.: {The
  extended Cohn-Kanade dataset (CK+)}: A complete dataset for action unit and
  emotion-specified expression. In: IEEE Computer Society Conference on
  Computer Vision and Pattern Recognition Workshops. pp. 94--101 (2010)

\bibitem{pantic2005web}
Pantic, M., Valstar, M., Rademaker, R., Maat, L.: Web-based database for facial
  expression analysis. In: IEEE International Conference on Multimedia and
  Expo. pp. 317--321 (2005)

\bibitem{phoo2020self}
Phoo, C.P., Hariharan, B.: Self-training for few-shot transfer across extreme
  task differences. In: International Conference on Learning Representations
  (2021)

\bibitem{ruan2020deep}
Ruan, D., Yan, Y., Chen, S., Xue, J.H., Wang, H.: Deep disturbance-disentangled
  learning for facial expression recognition. In: ACM International Conference
  on Multimedia. pp. 2833--2841 (2020)

\bibitem{ruan2021feature}
Ruan, D., Yan, Y., Lai, S., Chai, Z., Shen, C., Wang, H.: Feature decomposition
  and reconstruction learning for effective facial expression recognition. In:
  IEEE/CVF Conference on Computer Vision and Pattern Recognition. pp.
  7660--7669 (2021)

\bibitem{simon2020adaptive}
Simon, C., Koniusz, P., Nock, R., Harandi, M.: Adaptive subspaces for few-shot
  learning. In: IEEE/CVF Conference on Computer Vision and Pattern Recognition.
  pp. 4136--4145 (2020)

\bibitem{slimani2019compound}
Slimani, K., Lekdioui, K., Messoussi, R., Touahni, R.: Compound facial
  expression recognition based on highway {CNN}. In: New Challenges in Data
  Sciences: Acts of the Second Conference of the Moroccan Classification
  Society. pp.~1--7 (2019)

\bibitem{snell2017prototypical}
Snell, J., Swersky, K., Zemel, R.S.: Prototypical networks for few-shot
  learning. In: Advances in Neural Information Processing Systems. pp.
  4077--4087 (2017)

\bibitem{sung2018learning}
Sung, F., Yang, Y., Zhang, L., Xiang, T., Torr, P.H., Hospedales, T.M.:
  Learning to compare: Relation network for few-shot learning. In: IEEE/CVF
  Conference on Computer Vision and Pattern Recognition. pp. 1199--1208 (2018)

\bibitem{tian2020rethinking}
Tian, Y., Wang, Y., Krishnan, D., Tenenbaum, J.B., Isola, P.: Rethinking
  few-shot image classification: A good embedding is all you need? In: European
  Conference on Computer Vision. pp. 266--282 (2020)

\bibitem{vinyals2016matching}
Vinyals, O., Blundell, C., Lillicrap, T., Kavukcuoglu, K., Wierstra, D.:
  Matching networks for one shot learning. In: Advances in Neural Information
  Processing Systems. pp. 3630--3638 (2016)

\bibitem{wah2011caltech}
Wah, C., Branson, S., Welinder, P., Perona, P., Belongie, S.: The {Caltech-UCSD
  Birds-200-2011} dataset  (2011)

\bibitem{wang2019identity}
Wang, C., Wang, S., Liang, G.: Identity- and pose-robust facial expression
  recognition through adversarial feature learning. In: ACM International
  Conference on Multimedia. pp. 238--246 (2019)

\bibitem{wang2020suppressing}
Wang, K., Peng, X., Yang, J., Lu, S., Qiao, Y.: Suppressing uncertainties for
  large-scale facial expression recognition. In: IEEE/CVF Conference on
  Computer Vision and Pattern Recognition. pp. 6897--6906 (2020)

\bibitem{yang2021free}
Yang, S., Liu, L., Xu, M.: Free lunch for few-shot learning: Distribution
  calibration. In: International Conference on Learning Representations (2021)

\bibitem{zeng2018facial}
Zeng, J., Shan, S., Chen, X.: Facial expression recognition with inconsistently
  annotated datasets. In: European Conference on Computer Vision. pp. 222--237
  (2018)

\bibitem{zhang2020two}
Zhang, Z., Yi, M., Xu, J., Zhang, R., Shen, J.: Two-stage recognition and
  beyond for compound facial emotion recognition. In: IEEE International
  Conference on Automatic Face and Gesture Recognition. pp. 900--904 (2020)

\bibitem{zhao2018dynamic}
Zhao, F., Zhao, J., Yan, S., Feng, J.: Dynamic conditional networks for
  few-shot learning. In: European Conference on Computer Vision. pp. 19--35
  (2018)

\bibitem{zhao2011facial}
Zhao, G., Huang, X., Taini, M., Li, S.Z., Pietik{\"a}Inen, M.: Facial
  expression recognition from near-infrared videos. Image and Vision Computing
  \textbf{29}(9),  607--619 (2011)

\bibitem{zhou2021binocular}
Zhou, Z., Qiu, X., Xie, J., Wu, J., Zhang, C.: Binocular mutual learning for
  improving few-shot classification. In: IEEE/CVF International Conference on
  Computer Vision. pp. 8402--8411 (2021)

\bibitem{zhu2022convolutional}
Zhu, Q., Mao, Q., Jia, H., Noi, O.E.N., Tu, J.: Convolutional relation network
  for facial expression recognition in the wild with few-shot learning. Expert
  Systems with Applications  \textbf{189},  116046 (2022)

\bibitem{zou2022facial}
Zou, X., Yan, Y., Xue, J.H., Chen, S., Wang, H.: When facial expression
  recognition meets few-shot learning: A joint and alternate learning
  framework. In: AAAI Conference on Artificial Intelligence (2022)

\end{thebibliography}
\end{document}